\documentclass{article}

\usepackage{microtype}
\usepackage{graphicx}
\usepackage{subcaption}
\usepackage{booktabs} 

\usepackage{hyperref}

\usepackage{graphicx}
\newcommand\sbullet[1][.5]{\mathbin{\vcenter{\hbox{\scalebox{#1}{$\bullet$}}}}}

\usepackage[preprint]{icml2026}

\usepackage{amsmath}
\usepackage{amssymb}
\usepackage{mathtools}
\usepackage{amsthm}
\usepackage{comment}

\usepackage[most]{tcolorbox}
\usepackage{fancyvrb}
\usepackage{fvextra}
\DefineVerbatimEnvironment{MyVerbatim}{Verbatim}{breaklines, fontsize=\small}

\usepackage{hyperref}

\hypersetup{
  colorlinks=true,
  linkcolor=blue,
  citecolor=blue,
  urlcolor=blue
}
\usepackage{enumitem}
\usepackage{float}

\usepackage{longtable}
\usepackage{booktabs}

\usepackage[absolute,overlay]{textpos}
\usepackage[capitalize,noabbrev]{cleveref}
\usepackage{float}

\theoremstyle{plain}
\newtheorem{theorem}{Theorem}[section]
\newtheorem{proposition}[theorem]{Proposition}
\newtheorem{lemma}[theorem]{Lemma}
\newtheorem{corollary}[theorem]{Corollary}
\theoremstyle{definition}
\newtheorem{definition}[theorem]{Definition}
\newtheorem{assumption}[theorem]{Assumption}
\theoremstyle{remark}
\newtheorem{remark}[theorem]{Remark}

\icmltitlerunning{EchoReview:Learning Peer Review from the Echoes of Scientific Citations}

\begin{document}

\twocolumn[
  \icmltitle{EchoReview: Learning Peer Review from the Echoes of Scientific Citations \\
    }



  \icmlsetsymbol{equal}{*}

  \begin{icmlauthorlist}
  \icmlauthor{Yinuo Zhang}{hit}
  \icmlauthor{Dingcheng Huang}{hit}
  \icmlauthor{Haifeng Suo}{hit}
  \icmlauthor{Yizhuo Li}{hit}
  \icmlauthor{Ziya Zhao}{hit}
  \icmlauthor{Junhao Xu}{hit}
  \icmlauthor{Zhiying Tu}{hit}
  \icmlauthor{Dianhui Chu}{hit}
  \icmlauthor{Deming Zhai}{hit}
  \icmlauthor{Xianming Liu}{hit}
  \icmlauthor{Xiaoyan Yu}{bit}
  \icmlauthor{Dianbo Sui}{hit}
\end{icmlauthorlist}

\icmlaffiliation{hit}{Harbin Institute of Technology, China}
\icmlaffiliation{bit}{Beijing Institute of Technology, China}

\icmlcorrespondingauthor{Dianbo Sui}{suidianbo@hit.edu.cn}


  \vskip 0.3in
]
\printAffiliationsAndNotice{}




\begin{abstract}
As the volume of scientific submissions continues to grow rapidly, traditional peer review systems are facing unprecedented scalability pressures, highlighting the urgent need for automated reviewing methods that are both scalable and reliable. Existing supervised fine-tuning approaches based on real review data are fundamentally constrained by single-source of data as well as the inherent subjectivity and inconsistency of human reviews, limiting their ability to support high-quality automated reviewers. To address these issues, we propose EchoReview, a citation-context–driven data synthesis framework that systematically mines implicit collective evaluative signals from academic citations and transforms scientific community’s long-term judgments into structured review-style data. Based on this pipeline, we construct EchoReview-16K, the first large-scale, cross-conference, and cross-year citation-driven review dataset, and train an automated reviewer, EchoReviewer-7B. Experimental results demonstrate that EchoReviewer-7B can achieve significant and stable improvements on core review dimensions such as evidence support and review comprehensiveness, validating citation context as a robust and effective data paradigm for reliable automated peer review.
\end{abstract}

\section{Introduction}

Scientific research is fundamentally driven by the closed-loop mechanism of ``submission--review--revision,'' which is essential for ensuring academic quality and rigor~\citep{Smith2006PeerReview,Boughton2018ResearchIntegrity}. Peer review provides in-depth feedback that not only offers valuable insights to authors but also plays a critical role in improving the impact of research. However, in recent years, the number of submissions across disciplines has grown explosively. For example, AAAI-2026 received approximately 23,680 submissions, a sharp increase from 12,957 submissions in 2025, representing a year-over-year growth of 82.7\% and marking the highest submission volume in the conference's 40-year history. This rapid growth has far exceeded the capacity of traditional peer review systems~\citep{tran2020openreviewopenreviewcritical}, leading to excessive reviewer workload, review fatigue, superficial feedback, and insufficient argumentation, which collectively undermine review quality~\citep{wei2025aiimperativescalinghighquality,Adam2025ThePC}.

To bridge this volume gap, the automation of the academic review process has become imperative. Large language models (LLMs) have demonstrated transformative potential for fully automated peer review~\citep{Zhuang_2025}. One mainstream approach is supervised fine-tuning (SFT), which improves the performance and adaptability of pretrained models via domain-specific labeled data~\citep{han2024parameterefficientfinetuninglargemodels}. Prior work has attempted to construct large-scale real-world review datasets to fine-tune models for review generation, such as ReviewMT ~\citep{tan2024peer} and DeepReviewer~\citep{zhu2025deepreviewimprovingllmbasedpaper}. However, current methods face two fundamental challenges: (1) \textbf{Single-source of peer review data}. Only a very limited number of academic conferences, exemplified by ICLR, systematically release complete peer-review records for both accepted and rejected submissions. Moreover, these conferences are largely confined to the artificial intelligence community that uses the OpenReview platform. This narrow data provenance severely restricts cross-disciplinary generalization and constitutes a major bottleneck for automated peer review research. (2) \textbf{Lack of review consistency}. Human peer reviews are inherently subjective~\citep{Anderson2009ConferenceReviewing,kang-etal-2018-dataset}, often leading to substantial variability in evaluation outcomes. A prominent example is the NeurIPS 2014 review consistency experiment~\citep{Cortes2021InconsistencyIC}, in which a subset of submissions was randomly selected and independently assessed by two disjoint review committees. The results revealed striking discrepancies in both review scores and acceptance decisions for the same papers, with some papers strongly recommended for acceptance by one committee but rejected by the other. This large inter-reviewer variance underscores the intrinsic subjectivity and instability of traditional peer review under single-round and limited-sample conditions, suggesting that fine-tuning LLMs solely on real-world review data cannot ensure objective or consistent evaluation behavior.



To address these fundamental challenges, we propose a citation-driven framework for automated review data generation. In detail, we are motivated by the observation that academic citations explicitly reference prior work and implicitly convey evaluative signals: positive citations often indicate recognition and strengths, while negative citations suggest limitations or weaknesses. By systematically mining these implicit review signals, we can construct large-scale, cross-disciplinary review data that provide a solid foundation for automated peer review. Such a paradigm can clearly offer two key advantages: (1) \textbf{Breaking the single-source bottleneck of peer review data and enabling scalable extension to other scientific disciplines}. Unlike publicly released peer review records, which are highly restricted, our methods do not rely on the open-review policies of conferences or journals. Instead, labeled data can be automatically collected from large-scale scholarly databases such as Semantic Scholar and PubMed, enabling cross-domain and cross-disciplinary generalization and opening a new path toward scalable automated reviewing systems. 
(2) \textbf{Mitigating individual bias through collective intelligence}. Traditional reviews reflect the transient opinions of a small number of anonymous reviewers and are susceptible to randomness and personal preferences. In contrast, citation contexts emerge from the scientific community’s long-term and cumulative citation behaviors, aggregating collective judgments across researchers and time scales. Such community-level feedback effectively reduces subjectivity and uncertainty inherent in single-round peer review, 
enabling a more comprehensive evaluation of a paper’s long-term impact and improvement opportunities.


Based on these observations, we introduce EchoReview, an automated review data synthesis framework centered on citation context. EchoReview systematically transforms implicit evaluative signals embedded in long-term citation behaviors into structured and interpretable review-style data. Specifically, EchoReview introduces an end-to-end data generation pipeline: starting from high-impact papers, it automatically traces their subsequent citing works; extracts precise citation contexts and identifies positive and negative evaluative signals; refines, expands, and structures these implicit evaluations using LLMs; and finally incorporates evidence retrieval and multi-model cross-auditing to generate high-quality review samples with explicit justifications and chains of thought (CoT). This pipeline does not rely on human annotation or publicly available review records and can be stably scaled across conferences, years, and even disciplines, providing a new data paradigm for training large-scale automated reviewers. 

Armed with the pipeline, we construct \texttt{EchoReview-16K}, the first large-scale, cross-conference, and cross-year synthetic peer-review dataset derived from citation contexts, covering papers from ACL, EMNLP, ICLR, ICML, and NeurIPS between 2020 and 2022. Built upon this dataset, we further  introduce \texttt{EchoReview-Bench} to systematically evaluate the capabilities of different AI reviewers in realistic academic peer review settings and train \texttt{EchoReviewer-7B}. Experimental results demonstrate that EchoReviewer-7B achieves stable and significant improvements over existing models on core review dimensions, including evidence support and review comprehensiveness.

Our main contributions are summarized as follows:
\begin{itemize}[topsep=0pt, itemsep=3pt, parsep=0pt,leftmargin=2em]
  \item We propose a new paradigm for automatically synthesizing high-quality peer review data, independent of the open-review policies of conferences or journals.
  To this end, we systematically exploit the implicit collective evaluative signals embedded in citation contexts to build scalable training data, and develop an end-to-end automated review generation framework named EchoReview.
  
  \item We construct EchoReview-16K, the first large-scale citation-based review dataset. 
  This dataset contains structured Strengths and Weaknesses reviews along with their corresponding CoT rationales, providing a solid foundation for training high-quality AI reviewers.
  
  \item We train and comprehensively evaluate EchoReviewer-7B. By conducting multi-model comparisons and parallel analyses with human reviews, we systematically characterize the capability profiles of different AI reviewers. Our results highlight the unique advantages of citation-driven data in improving evidential support and review reliability.
\end{itemize}

\section{Methodology}
\label{sec:echoreview}



\begin{figure*}[t] 
    \centering
    \includegraphics[width=1.0\textwidth]{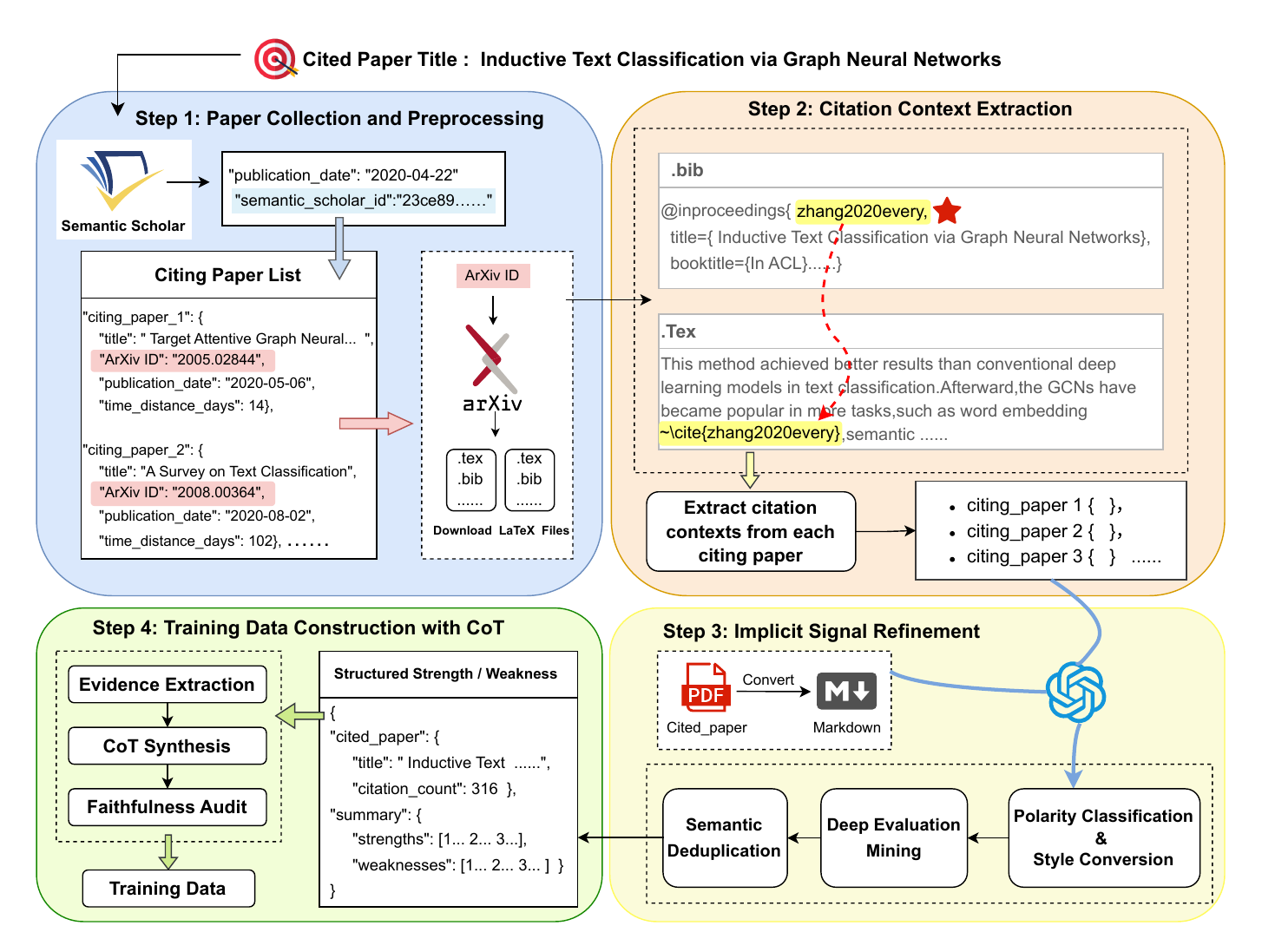} 
    \caption{Overview of the EchoReview automated data construction pipeline.}
    \label{fig:pipeline}
    \vspace{-10pt}

\end{figure*}

\subsection{Automated Data Construction in EchoReview}
\label{sec:pipeline}

To automatically construct high-quality review data, EchoReview focuses on cited papers and systematically extracts their citation contexts from subsequent studies, transforming implicit evaluation signals into review-style comments. The overall pipeline is illustrated in Figure~\ref{fig:pipeline} and consists of four stages: paper collection and preprocessing, citation context extraction, implicit signal refinement, and training data construction.

\subsubsection{Paper Collection and Preprocessing} To ensure data quality and information density, we prioritize highly influential papers with sufficient citations. Specifically, we collected all accepted papers from ACL, EMNLP, ICLR, ICML, and NeurIPS between 2020 and 2022 from the conference websites, forming the initial list of cited papers. 
\vspace{-2pt}
\begin{itemize}[topsep=0pt, itemsep=3pt, parsep=0pt,leftmargin=1em]
    \item \textbf{Metadata retrieval:} Based on paper titles, we query the Semantic Scholar and retain only exact title matches to avoid ambiguities caused by title changes. For each paper, we collect its unique Semantic Scholar identifier (paperId), publication date, and citation count for subsequent citation tracking and temporal filtering.
    \item \textbf{Citing papers selection:} For each cited paper, we retrieve its citing papers using the paperId and collect metadata including publication date and ArXiv identifier. Papers without ArXiv IDs are excluded. The citing papers are then sorted by temporal proximity to the cited paper. Using the ArXiv ID, we download the LaTeX source files including \texttt{.tex} and \texttt{.bib} files.
    \item \textbf{Cited paper format conversion:} We use the MinerU tool~\citep{wang2024mineruopensourcesolutionprecise} to convert cited paper PDFs into parseable Markdown format, removing author information, acknowledgments, references, and appendices, keeping only the main content to preserve technical details while avoiding unrelated information.
\end{itemize}
\subsubsection{Citation Context Extraction.} This stage can establish precise mappings between cited and citing papers via the following steps:
\vspace{-2pt}
\begin{itemize}[topsep=0pt, itemsep=3pt, parsep=0pt,leftmargin=1em]
    \item \textbf{BibKey anchoring:} Using the cited paper title, we locate the corresponding \texttt{bib\_key} in the citing paper's \texttt{.bib} file.
    \item \textbf{Context window extraction:} We scan the \texttt{.tex} source of the citing paper, using regular expressions to locate all occurrences of \texttt{\textbackslash cite\{bib\_key\}}. A three-sentence sliding window (the target sentence and its preceding and following sentences) is used to extract the citation context, preserving semantic polarity and argumentative logic.
\end{itemize}

\subsubsection{Implicit Signal Enhancement}
\label{sec:Implicit Signal Enhancement}

Since citation contexts are often fragmented and contain many neutral statements, we use GPT-4o~\citep{openai2024gpt4ocard} to process and structure them via the following steps\footnote{The complete prompts are provided in the Appendix~\ref{sec:1}}:
\vspace{-2pt}
\begin{itemize}[topsep=0pt, itemsep=3pt, parsep=0pt,leftmargin=1em]
    \item \textbf{Polarity classification and style conversion:} We leverage the LLM to classify citation polarity (Strength/Weakness/Neutral), retaining only positive and negative citations, which are then transformed into standardized review comments.
    \item \textbf{Deep evaluation mining:} To capture implicit evaluation signals, we design diagnostic questions to identify indirect endorsements or potential weaknesses, such as method adoption, experimental setup replication, or evaluation standard inheritance. Model responses are integrated into the corresponding Strength or Weakness comments (The full list of questions is provided in Appendix~\ref{sec:1.2}.)
    \item \textbf{Semantic deduplication: } Since different citation contexts can provide highly similar evaluative statements around the same contribution or limitation, we apply a semantic-level deduplication process to the generated review comments. Rather than relying on surface-level lexical overlap, we assess duplication at the semantic level by determining whether two comments point to the same core evaluative insight. For each group of semantically redundant comments, we retain the version that is the most comprehensive, clearly articulated, and evaluatively informative, while removing the others. The retained comment preserves its original phrasing without any rewriting. Through this process, each review sample ultimately contains a set of semantically non-overlapping key evaluation points, improving information density and preventing redundant signals from interfering with subsequent model training.
\end{itemize}

\subsubsection{Training Data Construction}
\label{Training Data Construction with CoT}

Although review comments generated from citation contexts can effectively capture evaluative signals from the scholarly community, such comments are often limited to high-level conclusions, lacking explicit, traceable supporting evidence and intermediate reasoning. This limitation can weaken both the logical rigor and interpretability of automated reviewer models during training. To address this issue, we augment the EchoReview framework by incorporating a CoT structure based on the \emph{Evidence-Reasoning-Conclusion} format\footnote{The complete prompts are provided in the Appendix~\ref{sec:2}.}:
\vspace{-10pt}
\begin{itemize}[topsep=0pt, itemsep=3pt, parsep=0pt, leftmargin=1em]
    \item \textbf{Evidence Extraction:}  
    Given a cited paper and its associated review comment, we first retrieve textual evidence from the full paper that can directly support the corresponding evaluative judgment. This step is performed by GPT-4o, whose goal is to locate 1--3 supporting evidence passages that can serve as explicit justifications for the review comment. All extracted evidence must be verbatim excerpts from the original paper, rather than model-generated paraphrases or summaries. In addition, each evidence passage is accompanied by a brief explanation clarifying how it supports the corresponding Strength or Weakness judgment. Through this explicit evidence alignment mechanism, we ensure that every review comment can be traced back to verifiable textual sources, thereby avoiding unsupported claims or hallucinated evaluations.

    \item \textbf{CoT Synthesis:}  In this step, the textual evidence extracted during the Evidence Extraction phase, along with the full paper content, is fed into GPT-4o. GPT-4o is then tasked with generating a compact, evidence-centered Chain-of-Thought, where each reasoning step starts with direct quotations from the original paper, followed by a few analytical sentences explaining the logical connection between the evidence and the evaluative judgment, and concludes with a clear and concrete statement of Strength or Weakness.

\begin{figure}[t]
    \centering
    \includegraphics[width=0.9\linewidth]{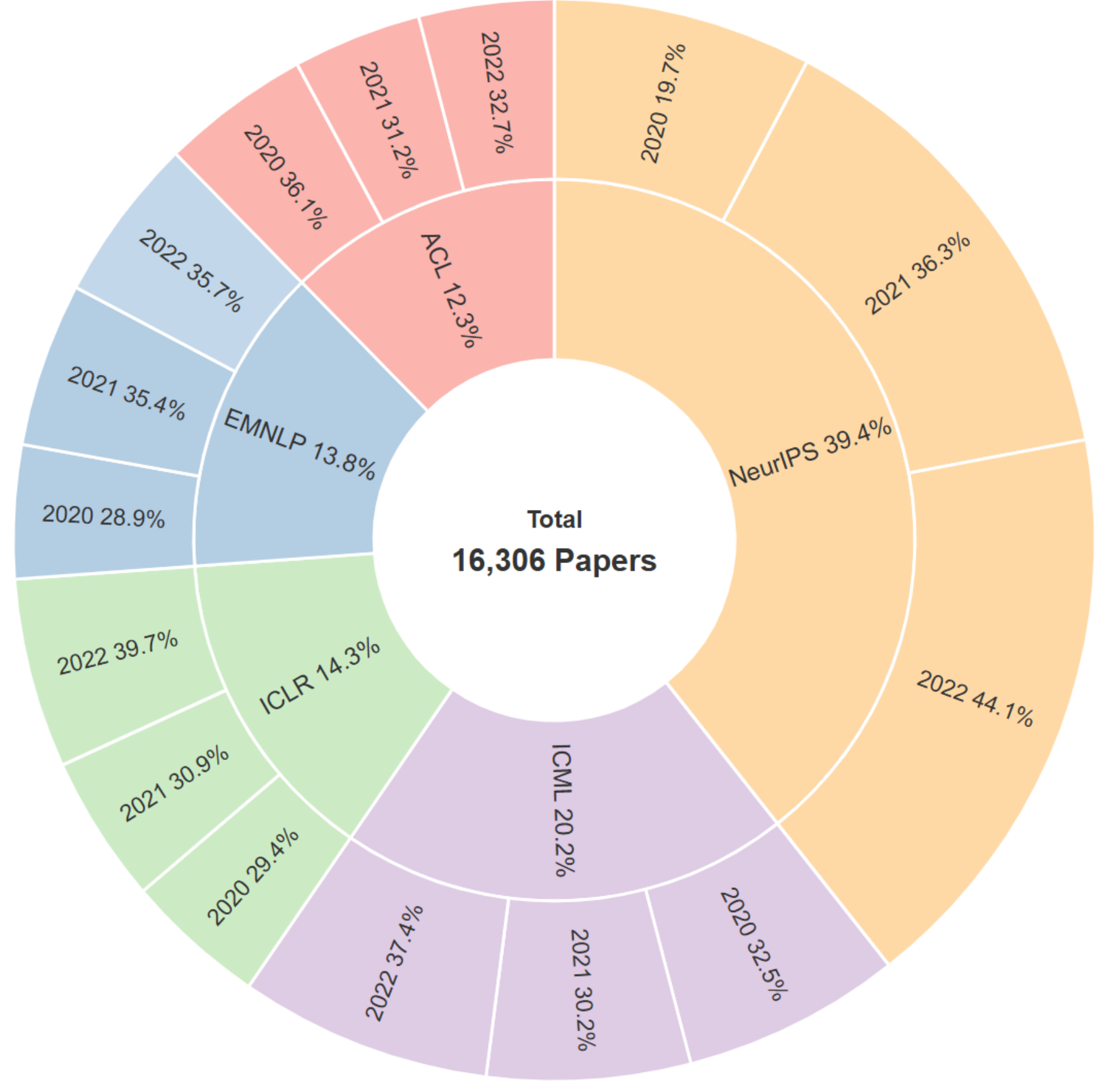}
    \caption{Statistical distribution of the EchoReview-16K dataset. The dataset comprises 16,306 high-quality samples collected from five top-tier AI conferences (ACL, EMNLP, ICLR, ICML, and NeurIPS) across three consecutive years (2020--2022).}
    \label{fig:statistics}
\vspace{-12pt}
\end{figure}

    \item \textbf{Multi-model Cross-validation (Faithfulness Audit):}  
    To further ensure the faithfulness and logical consistency of the generated CoTs, we introduce Qwen-max~\cite{qwen25} as an independent auditing model to verify the reasoning process. The auditor evaluates each CoT based on the original evidence passages and the full paper text along three dimensions: (1) \emph{citation validity}, assessing whether all claims in the reasoning are supported by the cited evidence; (2) \emph{logical coherence}, examining whether the reasoning contains leaps, contradictions, or inconsistencies; and (3) \emph{overall explanatory quality}, measuring whether the CoT clearly and reasonably maps evidence to conclusions. The audit model outputs structured scores along with deduction rationales, and only samples that pass all checks and exceed a predefined quality threshold are retained.

    \item \textbf{Sample Packaging:}  
    Finally, the system prompt and the full text of the cited paper are used as inputs, while the structured Strength/Weakness comments augmented with Chain-of-Thought are used as outputs to construct training samples for supervised fine-tuning (SFT).
\end{itemize}

\subsection{EchoReview-16K and EchoReview-Bench}
\label{sec:dataset}

Based on the pipeline described above, we construct 16,306 high-quality review samples from ACL, EMNLP, ICLR, ICML, and NeurIPS papers published between 2020 and 2022, forming the EchoReview-16K dataset. It covers multiple research directions and includes structured Strength/Weakness feedback with corresponding reasoning chains. Dataset statistics are shown in Figure~\ref{fig:statistics}.

To systematically evaluate the quality of AI-generated reviews, we further construct EchoReview-Bench. We randomly sampled 1,631 examples (about 10\%) from EchoReview-16K and divided them into two complementary test sets:
\vspace{-2pt}
\begin{itemize}[topsep=0em, itemsep=3pt, parsep=0pt, leftmargin=1em]

    \item \textbf{Test Set A (General Review Evaluation):} Contains 1,398 papers from ACL, EMNLP, ICML, and NeurIPS.
    \item \textbf{Test Set B (Human Reviewer Reference):} Contains only ICLR papers. From 2,330 ICLR papers published 2020–2022, we randomly sampled 233 papers (about 10\%) and collected all corresponding human review comments from OpenReview.
\end{itemize}
\vspace{-2pt}
\subsection{EchoReviewer-7B}
We fine-tune Qwen2.5-7B-Instruct on the training set of EchoReview-Bench using LoRA~\citep{hu2021loralowrankadaptationlarge}. Training is performed on two NVIDIA RTX A6000 GPUs (48GB) via the LLaMA-Factory framework~\citep{zheng2024llamafactory}.
\section{Experiments}
\label{sec:Exp}
\subsection{Overall Review Quality Evaluation}
\label{sec:main_exp}

This experiment aims to systematically evaluate the overall review quality of EchoReviewer in realistic scenarios and compare it with existing representative AI reviewers.

\noindent\textbf{Setup.}  
The main experiment is conducted on Test Set A of EchoReview-Bench, which can adequately reflect the model's capability in general academic peer review scenarios.  

Inspired by prior work, such as TreeReview~\citep{chang2025treereviewdynamictreequestions}, we adopt an LLM-as-Judge automatic evaluation paradigm, using Gemini-2.5-Pro~\citep{comanici2025gemini25pushingfrontier} as the judge model to score reviews generated by different systems across multiple quality dimensions. Specifically, the evaluation covers the following four key dimensions:  
\vspace{-2pt}
\begin{itemize}[topsep=0pt, itemsep=3pt, parsep=0pt,leftmargin=1em]
    \item \textbf{Comprehensiveness:} Whether the review systematically covers the paper's main contributions, methods, experiments, and potential issues.
    \item \textbf{Specificity:} Whether the review provides concrete, actionable feedback rather than generic statements.
    \item \textbf{Evidence Support:} Whether the review points are backed by explicit paper content or experimental results.
    \item \textbf{Consistency:} Whether the review is internally coherent and logically self-consistent.
\end{itemize}

The judge model assigns a score from 0 to 10 for each dimension and computes the mean across all four dimensions as the Overall Quality score. To enhance evaluation stability and robustness, each review is independently scored three times under a temperature of 0.1, and the average is taken as the final score. The judge model is also required to provide concise textual explanations for each dimension and return results in a structured JSON format. Complete evaluation prompts are provided in the Appendix~\ref{sec:3.1}.  

\textbf{Baselines.}  
We select two categories of representative reviewer models for comparison: 
\vspace{-2pt}
\begin{enumerate}[topsep=0pt, itemsep=3pt, parsep=0pt,leftmargin=1em]
    \item \textbf{Prompt-based reviewer models}, including Claude Opus 4.5~\citep{claude45}, DeepSeek-R1~\citep{DS-R1}, Gemini 3~\citep{google_gemini3}, and GPT-5~\citep{gpt5card}. All these LLMs use a unified prompt in the experiments, which is provided in the Appendix~\ref{sec:3.2}.
    \item \textbf{Fine-tuned reviewer models}, including CycleReviewer-8B~\citep{cycleresearcher} and DeepReviewer-7B~\citep{zhu2025deepreviewimprovingllmbasedpaper}.
\end{enumerate}

\begin{figure}[t]
    \centering
    \includegraphics[width=\linewidth]{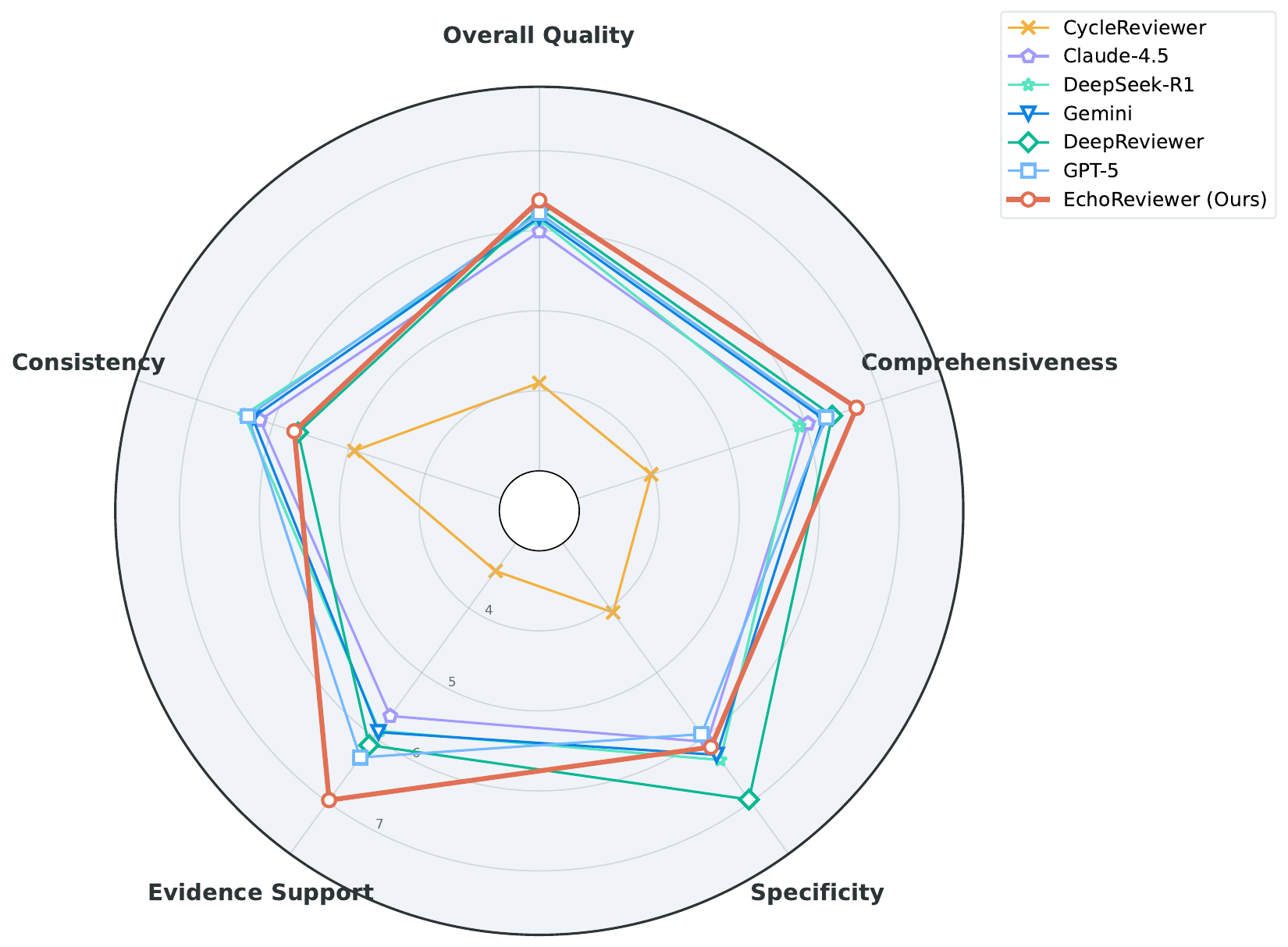} 
    \caption{Performance comparison of reviewer models on multiple quality dimensions.}
    \label{fig:radar_chart}
    \vspace{-12pt}
\end{figure}

\textbf{Results.}  
Experimental results are shown in Figure~\ref{fig:radar_chart}. The results indicate that the performance advantages of EchoReviewer-7B are consistently reflected in several core dimensions critical to automated peer review. Notably, EchoReviewer attains the highest score among all methods in the Evidence Support dimension, demonstrating that it tends to provide verifiable arguments grounded in paper content rather than relying on generalized evaluation language. This characteristic is particularly important in real academic review scenarios, where feedback lacking evidence often fails to provide actionable guidance for authors. Similarly, EchoReviewer leads in Comprehensiveness, showing that it can systematically cover a paper’s contributions and potential shortcomings from multiple perspectives. Compared with models trained only on single-round human review data or general instruction fine-tuning, these results reflect the advantage of citation-driven training in integrating long-term, collective evaluative signals from the academic community.  

In contrast, some general-purpose LLMs with higher overall scores demonstrate more stable language fluency and consistency but tend to lack sufficient evidence density and methodological specificity in their reviews. It should be noted that EchoReviewer still lags behind some general-purpose models in Consistency, indicating that citation-driven data construction may present structural challenges in global coherence. Citation contexts are inherently fragmented and local; although dense in information, integrating them into a coherent review may place higher demands on the model’s document planning capability. Therefore, the current results should not be interpreted as simple performance deficiencies but rather as differences in review capability emphasis arising from different data paradigms.  


\subsection{Comparison between Human Reviews and Citation-based Long-term Feedback}
\label{sec:human_vs_longterm}

Although recent studies have shown that AI reviewers can generate well-structured and fluent review reports, the prevailing consensus in the research community is that AI reviewers are better positioned as assistive tools rather than full replacements for human reviewers~\citep{xu2025llmsidentifycriticallimitations,zhu2025deepreviewimprovingllmbasedpaper}. Despite well-known issues such as subjectivity and limited inter-reviewer consistency, human peer review remains a central and irreplaceable component of the academic evaluation system. Accordingly, the goal of this experiment is neither to assess whether EchoReviewer is ``better than'' human reviewers, nor to measure their degree of alignment, but rather to \textbf{investigate the \underline{complementary relationship} between the two in terms of review perspectives and the types of issues they emphasize}.



\noindent\textbf{Setup.}
This experiment is conducted on Test Set B of EchoReview-Bench, which consists of 233 papers published at ICLR between 2020 and 2022. For each paper, we collect all corresponding human reviews from OpenReview and extract the weaknesses identified by the reviewers. Under identical full-paper input conditions, we further generate review reports using EchoReviewer and three general-purpose large language models (DeepSeek-R1, Gemini~3, and GPT-5), from which the Weakness sections are extracted.

Since multiple weakness statements may refer to the same underlying research limitation, we do not treat individual weakness statements as the unit of analysis. Instead, we analyze \textit{underlying research issues}. An underlying research issue refers to a fundamental limitation of a paper at the level of methodological assumptions, problem formulation, experimental design, or applicability, rather than specific surface-level phrasing or implementation details. Given the substantial variation in expression and focus across different sources, we do not require strict textual equivalence. Instead, two weakness statements are considered matched if they point to the same underlying research issue. This consistency judgment is performed by GPT-5.2~\citep{openai_gpt52_2025}, which is used solely to determine whether two weakness statements refer to the same underlying research issue and outputs a binary decision (with temperature fixed to 0). GPT-5.2 does not generate new weaknesses and does not participate in any quality assessment; it serves exclusively as a consistency judgment tool. All prompts used during evaluation are provided in the Appendix~\ref{sec:3.3}.

\noindent\textbf{Metrics.}
We compute three ratio-based metrics at the level of underlying research issues. In detail, for a given paper, let the set of research issues identified by human reviewers be $\mathcal{H} = \{h_1, h_2, \dots, h_n\}$.  For a given AI reviewer model $M$, the set of research issues it identifies is denoted as: $\mathcal{E}^{(M)} = \{e_1^{(M)}, e_2^{(M)}, \dots, e_m^{(M)}\}$.  The intersection between human reviewers and model $M$ at the research-issue level is defined as
\begin{equation}
\small
\mathcal{O}^{(M)} = \mathcal{H} \cap \mathcal{E}^{(M)}.
\end{equation}
Based on these sets, we can define three ratio-based metrics:
$\sbullet[.75]$ \textbf{Overlap Ratio} (Shared research issues):
    \begin{equation}
    \small
    R_{\text{overlap}}^{(M)} =
    \frac{|\mathcal{O}^{(M)}|}{|\mathcal{H} \cup \mathcal{E}^{(M)}|}.
    \end{equation}
$\sbullet[.75]$ \textbf{Human-only Ratio} (Human-only research issues):
    \begin{equation}
    \small
    R_{\text{human-only}}^{(M)} =
    \frac{|\mathcal{H} \setminus \mathcal{O}^{(M)}|}{|\mathcal{H}|}.
    \end{equation}
$\sbullet[.75]$ \textbf{Model-only Ratio} (Model-only research issues):
    \begin{equation}
    \small
    R_{\text{model-only}}^{(M)} =
    \frac{|\mathcal{E}^{(M)} \setminus \mathcal{O}^{(M)}|}{|\mathcal{E}^{(M)}|}.
    \end{equation}
All three metrics are first computed independently for each paper and then macro-averaged across the 233 papers in Test Set B to obtain the final reported results.

\begin{table}[t]
\centering
\caption{Comparison of underlying research issues identified by human reviewers and different AI reviewers.}
\label{tab:human_vs_model}
\begin{tabular}{lccc}
\toprule
\textbf{Model} &
$\boldsymbol{R_{\text{overlap}}^{(M)}}$ &
$\boldsymbol{R_{\text{human\text{-}only}}^{(M)}}$ &
$\boldsymbol{R_{\text{model\text{-}only}}^{(M)}}$ \\
\midrule
EchoReviewer-7B   & 0.252 & 0.195 & 0.553 \\
GPT-5          & 0.275 & 0.260 & 0.465 \\
DeepSeek-R1    & 0.341 & 0.275 & 0.384 \\
Gemini~3       & 0.382 & 0.354 & 0.264 \\
\bottomrule
\end{tabular}
\vspace{-12pt}
\end{table}
\noindent\textbf{Results.}
The experimental results are summarized in Table~\ref{tab:human_vs_model}. These results indicate that human reviews and citation-based automated reviews focus on different types of research issues and form a complementary coverage over the space of potential weaknesses. We also find that the research limitations identified by EchoReviewer are largely derived from usage-driven feedback that repeatedly appears in citation contexts. Such issues typically entail high discovery costs and often require validation and analysis beyond the scope of standard review cycles and available review resources, making them difficult to systematically identify within a short peer review period. Examples include scalability in large-scale or real-world deployment scenarios, robustness under distribution shifts, and numerical stability issues during training, which often only become apparent through subsequent reproduction and extension efforts.

It is important to clarify that the relatively high Model-only Ratio exhibited by EchoReviewer does not indicate a deviation from the peer review task. Rather, it suggests that \textbf{EchoReviewer can provide additional, forward-looking risk signals to human reviewers}. These research issues, distilled from long-term citation feedback, can be viewed as early indicators of potential failure modes, enabling reviewers to more comprehensively assess a method's reliability and applicability boundaries in real-world usage.

Meanwhile, general-purpose LLMs exhibit Overlap Ratio and Human-only Ratio values that are overall closer to those of human reviewers, indicating that the weaknesses they generate are more aligned with traditional submission-stage review conventions. However, they remain relatively conservative in uncovering latent research limitations that only emerge through long-term use. The structural differences in issue coverage between EchoReviewer-7B and general-purpose LLMs further demonstrate that incorporating citation-based long-term feedback provides a complementary and valuable source of information for peer review.


\subsection{Impact of Citation Time Span on Review Quality}


Since the core assumption of EchoReview is to leverage long-term collective evaluation signals from the academic community to mitigate the subjectivity inherent in single-round peer review, citation feedback at different temporal scales can exhibit systematic differences in information density and evaluative focus. Therefore, we further explore whether restricting the citation time window alters the performance characteristics of AI reviewers across different review dimensions.

\textbf{Setup.}
Based on EchoReview-16K, we retain only Strength and Weakness review samples whose citing papers were published within 500 days after the cited paper. A new model, \textit{EchoReviewer-Time-7B} (\textbf{TimeFilter}), is trained on this temporally filtered subset. Except for the citation time filtering, all model architectures and training settings remain identical to those of \textit{EchoReviewer-7B} (\textbf{NoTimeFilter}).

For evaluation, we continue to use Test Set A of EchoReview-Bench and adopt the same evaluation dimensions in Section~\ref{sec:main_exp}. Different from previous experiments, this study employs a \textit{pairwise battle} evaluation paradigm, directly comparing TimeFilter and NoTimeFilter.  To mitigate positional bias, the presentation order of the two candidate reviews is randomly permuted before being fed into the judge LLM. After reading the original paper and both reviews, the judge determines the better review along four dimensions: Comprehensiveness, Specificity, Evidence Support, and Consistency, and additionally provides an Overall Preference, reflecting the overall usefulness of the review.

%

\begin{figure}[t]
    \centering
    \includegraphics[width=\linewidth]{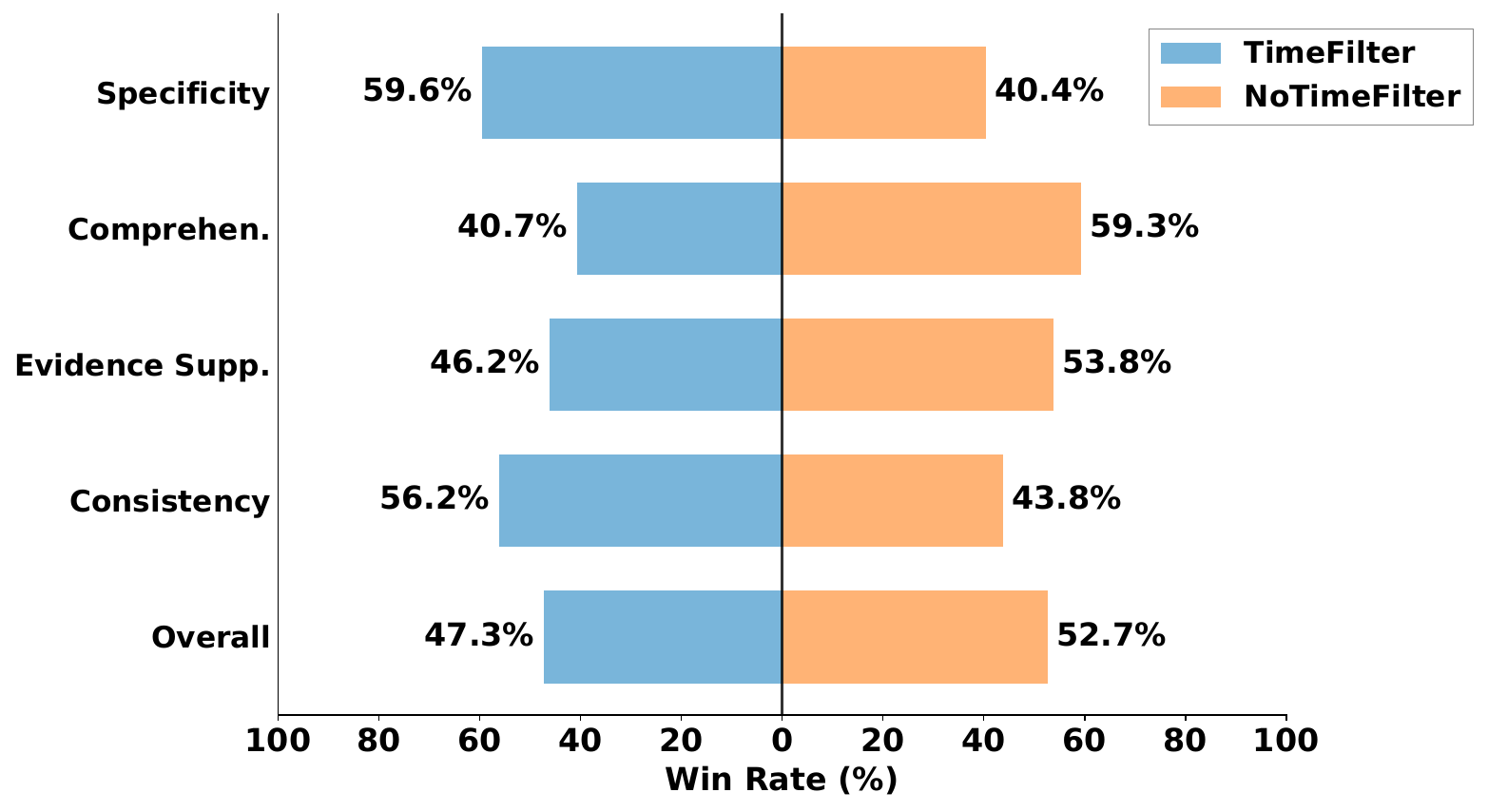}
    \caption{Pairwise comparison results between TimeFilter and NoTimeFilter across different review dimensions.}
    \label{fig:time_comparison}
    \vspace{-13pt}
\end{figure}

\noindent\textbf{Results.}
The results are summarized in Figure~\ref{fig:time_comparison}. The results reveal a clear trade-off between evaluation depth and evaluation breadth with respect to citation time span. In terms of overall preference, NoTimeFilter slightly outperforms TimeFilter, achieving a win rate of 52.7\%, suggesting that models trained on longer citation horizons tend to be more broadly acceptable when considering overall review usefulness.


However, a more fine-grained analysis across individual dimensions reveals pronounced divergences in review characteristics. NoTimeFilter demonstrates clear advantages in evaluation breadth and evidential robustness, with higher win rates in Comprehensiveness (59.3\%) and Evidence Support (53.8\%). This indicates that long-term citation corpora capture repeatedly validated experiential feedback from subsequent studies while introducing more diverse perspectives, enabling the model to assess a paper’s long-term value and potential impact from a broader contextual viewpoint.

In contrast, expanding the citation time span also introduces heterogeneous and dispersed evaluative signals, which can adversely affect technical focus and internal coherence. As shown in the results, TimeFilter exhibits significant advantages in Consistency (56.2\%) and Specificity (59.6\%). This suggests that early-stage citation feedback is typically concentrated on core technical aspects such as implementation details, methodological soundness, and experimental design, leading to a more coherent evaluative context. By restricting the citation time window, TimeFilter generates more targeted and tightly reasoned technical reviews, thereby avoiding viewpoint dispersion or internal conflicts caused by absorbing excessive long-term, cross-domain evaluative signals.
\begin{figure}[t]
    \centering
    \includegraphics[width=\linewidth]{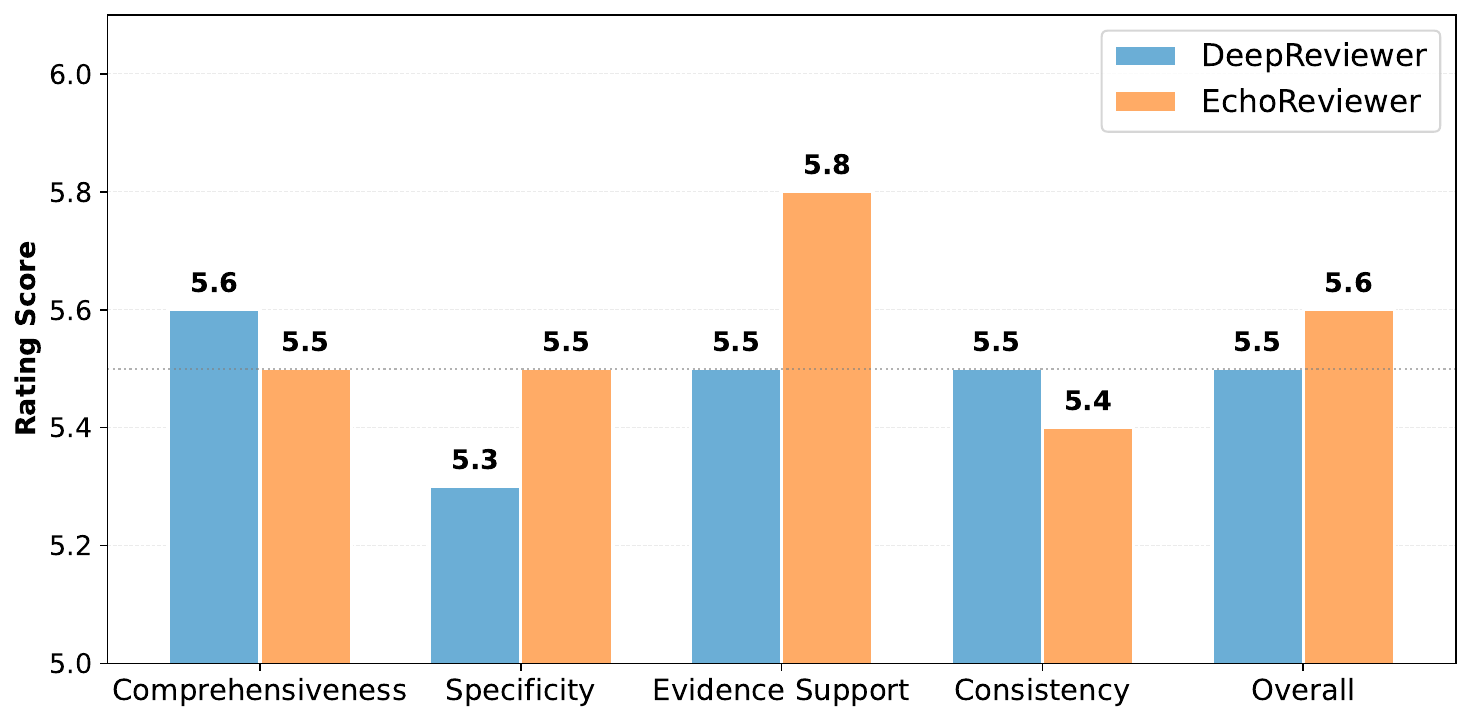}
    \caption{Comparison of review quality across five evaluation dimensions between DeepReviewer and EchoReviewer after incremental fine-tuning. }
    \label{fig:synergy_analysis}
    \vspace{-10pt}
\end{figure}
\subsection{Synergy Analysis: Cross-Paradigm Knowledge Integration}
\label{sec:synergy}

In Section~\ref{sec:main_exp}, we observe that EchoReviewer trained on citation contexts exhibits a natural advantage in Evidence Support, whereas DeepReviewer, trained on human-written reviews, performs better in review granularity and actionability, reflected by higher Specificity. These two paradigms correspond to complementary dimensions of academic evaluation: EchoReview-16K captures the long-term collective wisdom of the research community, while DeepReview-13K more closely reflects expert reviewers' procedural expertise and structured reasoning patterns. Based on these observations, we further investigate the potential synergy between the two paradigms. Specifically, we examine whether incremental fine-tuning can integrate expert-level structured reasoning into a citation-aware model backbone, thereby systematically improving the overall quality and robustness of automated peer review.

\noindent\textbf{Setup.}
We adopt the proposed EchoReviewer-7B as the base model and perform incremental SFT on the DeepReview-13K dataset.  For evaluation, we randomly sample 200 papers from Test Set A in EchoReview-Bench. To eliminate potential confounding effects from review aggregation, the reviewer number for DeepReviewer is fixed to one. We follow the LLM-as-Judge evaluation protocol described in Section~\ref{sec:main_exp}, and the evaluation is conducted along five core dimensions: Comprehensiveness, Specificity, Evidence Support, Consistency, and Overall Quality. The overall score is computed as the average of the first four dimensions, with all scores ranging from 0 to 10.

\noindent\textbf{Results.}
The quantitative results are illustrated in Figure~\ref{fig:synergy_analysis}. Overall, after the second-stage incremental fine-tuning, EchoReviewer demonstrates a more balanced and stable performance across all five evaluation dimensions, achieving an overall quality score of 5.6, which represents a +0.1 improvement over DeepReviewer. Although the absolute gain is modest, under a unified evaluation protocol and randomized sampling, this consistent improvement indicates a clear positive effect of synergistic training on overall review quality. More pronounced gains are observed on key capability dimensions. In particular, EchoReviewer achieves the highest score of 5.8 on {Evidence Support, substantially outperforming DeepReviewer (5.5). This suggests that the fact-anchoring ability induced by citation-context pretraining is not only preserved but further strengthened after incorporating expert-level reasoning data. Meanwhile, on Specificity—a dimension typically dominated by human-written review data—the model exhibits a stable improvement from 5.3 to 5.5. This indicates that expert structured reasoning can be effectively integrated with citation-driven evaluation signals, rather than causing negative transfer.

Taken together, these results suggest that using citation-driven collective consensus as the model backbone provides a solid foundation for factual grounding, while subsequent injection of human expert data further guides the model to map high-level citation signals onto concrete methodological details and technical diagnostics. This two-stage fine-tuning paradigm achieves a favorable balance between evaluative objectivity and diagnostic depth without introducing noticeable degradation, offering a practically valuable pathway toward building reliable automated peer review systems.

\section{Conclusion}
This paper proposes EchoReview, a citation-context--driven framework for automated peer review that constructs scalable and evidence-supported review data by mining implicit long-term collective evaluative signals embedded in scientific citations. We build the first cross-conference and cross-year citation-driven peer review dataset, EchoReview-16K, and train an automated reviewer, EchoReviewer-7B. Experimental results show that EchoReviewer-7B can achieve stable improvements on core review dimensions such as evidence support and review comprehensiveness. Moreover, citation-driven reviews are able to reveal long-term research limitations that differ from submission-stage human reviews and are oriented toward real-world usage and reproducibility. Overall, our findings show that citation context constitutes an effective data paradigm for automated peer review, serving as a valuable complement to human evaluation.

\section*{Impact Statement}
The development of EchoReview, a citation-driven framework for automated peer review data synthesis and AI reviewer training, holds substantial potential for improving the scalability, consistency, and evidence grounding of scholarly peer review. By transforming long-term community citation behaviors into structured review-style data, EchoReview aims to alleviate the structural scarcity of high-quality review datasets and reduce the workload burden faced by human reviewers. Nevertheless, automating components of the academic evaluation pipeline inherently raises important ethical, epistemic, and social considerations that require careful examination.

A primary concern lies in the risk of bias inheritance and amplification. Although citation contexts aggregate collective scientific judgments across time and communities, citation practices themselves are known to reflect systemic biases related to venue prestige, institutional reputation, geographic distribution, language, gender, and research popularity. As a result, models trained on citation-derived signals may inadvertently reinforce existing structural inequalities, potentially disadvantaging work from underrepresented groups, emerging institutions, novel methodologies, or interdisciplinary domains that traditionally receive fewer citations. Moreover, citation behavior is influenced not only by scientific merit but also by social dynamics and academic conventions, which may embed latent biases into the synthesized review data. Without careful design and auditing, EchoReview could thus propagate and even magnify these distortions.

Another important consideration concerns epistemic homogenization and reduced diversity of scientific perspectives. By learning from aggregated long-term citation patterns, AI reviewers may converge toward dominant paradigms and mainstream evaluation criteria, potentially undervaluing unconventional ideas, high-risk research, or paradigm-shifting contributions that initially receive limited recognition. Such effects could unintentionally narrow the intellectual diversity of scientific discourse and discourage exploratory or disruptive research. In addition, excessive reliance on automated reviewing tools risks deskilling human reviewers, diminishing critical engagement, and fostering overdependence on AI-generated judgments, which could ultimately weaken the rigor and accountability of the peer review process.

To proactively address these concerns, we adopt a multi-layered mitigation strategy throughout the design, training, and deployment of EchoReview. First, we carefully construct the EchoReview-16K dataset to ensure diversity across venues, years, and subfields, and we incorporate explicit positive and negative evaluative signals to prevent overly optimistic or one-sided assessments. We further employ multi-model cross-auditing and evidence retrieval mechanisms to enhance factual grounding and reduce hallucination and spurious reasoning. Second, we explicitly position EchoReviewer-7B as a decision-support system rather than an autonomous evaluator, strongly advocating a human-in-the-loop framework in which expert reviewers critically assess, contextualize, and override AI-generated feedback when necessary. We emphasize that final review decisions should always remain under human responsibility.

To promote transparency and accountability, we commit to open-sourcing our models, datasets, and pipelines, enabling community scrutiny, bias auditing, and reproducibility. Alongside code and data release, we will provide detailed documentation and usage guidelines that explicitly caution against blind reliance on automated outputs and highlight known limitations and failure modes. We also plan to conduct continuous bias benchmarking and domain-shift evaluation, systematically examining performance across disciplines, methodologies, and research communities to identify uneven behaviors and potential risks.

We believe that, when developed and deployed responsibly, EchoReview can serve as a valuable tool for improving the efficiency, fairness, and evidence quality of scholarly peer review. However, realizing this potential requires sustained community oversight, transparent reporting, and cautious integration into existing academic workflows. By embedding these principles into our design and release practices, we aim to maximize the societal benefits of EchoReview while minimizing its ethical and scientific risks.
\bibliography{example_paper}
\bibliographystyle{icml2026}

\newpage
\appendix
\onecolumn
\section{Related Work}

Automated Scientific Peer Review (ASPR) aims to leverage large language models (LLMs) to generate review reports, which constitute the core output of both traditional peer review and automated review systems. Existing ASPR approaches can be broadly categorized into three types: prompt-engineering-based methods, supervised fine-tuning-based methods, and multi-agent frameworks.

\begin{itemize}[leftmargin=1.5em]
    \item \textbf{Prompt-engineering-based methods.} These methods guide LLMs to generate structured and criteria-aligned review reports through carefully designed prompts~\citep{robertson2023gpt4slightlyhelpfulpeerreview,liang2023largelanguagemodelsprovide,Thelwall_2024,du2024llmsassistnlpresearchers,liu2023reviewergptexploratorystudyusing,zhou-etal-2024-llm}. Such approaches demonstrate the feasibility of using LLMs to produce coherent and format-compliant reviews; however, their performance is highly sensitive to prompt design and remains constrained by the models’ static knowledge and inherent capability limits.
    
    \item \textbf{Supervised fine-tuning-based methods.} This line of work improves review quality and consistency by constructing human-written review datasets and fine-tuning LLMs on them. Representative examples include ReviewMT, which models the review process as multi-turn dialogues ~\citep{tan2024peerreviewmultiturnlongcontext}, REVIEWER2, which adopts a stage-wise generation strategy ~\citep{gao2024reviewer2optimizingreviewgeneration}, and LimGen, which focuses on limitation generation for research papers~\citep{faizullah2024limgenprobingllmsgenerating}. Nevertheless, these methods heavily rely on human-authored review data and inevitably suffer from scalability limitations.
    
    \item \textbf{Multi-agent frameworks.} These approaches introduce multiple LLM agents with assigned roles to simulate the collaborative nature of real conference peer review, thereby generating more specific and comprehensive feedback ~\citep{jin2024agentreviewexploringpeerreview,chamoun2024automatedfocusedfeedbackgeneration,yu2024automatedpeerreviewingpaper,darcy2024margmultiagentreviewgeneration,wang2024mamorx}. Although multi-agent frameworks improve structural organization and content coverage, their review signals are still primarily derived from short-term, submission-stage judgments, lacking explicit modeling of long-term academic impact and real-world usage feedback.
\end{itemize}

\section{Prompt Details}
\label{sec:prompt-details}

Here are the prompts used in this paper.

\subsection{Prompts used in Implicit Signal Enhancement}
\label{sec:1}

\subsubsection{System Prompt}
\label{sec:1.1}
\begin{tcolorbox}[colback=yellow!10, colframe=black!50, title=Implicit Signal Enhancement -- System Prompt, breakable]
\begin{MyVerbatim}
You are a strict, objective AI Reviewer evaluating the "Cited Paper".
Your goal is to extract a peer-review assessment based solely on how other researchers (citations) have evaluated it.

**CORE PHILOSOPHY:**
You do NOT care about the "Citing Paper's" contribution. You only care about the "Cited Paper's" quality.
The citations are merely **evidence** or **testimony** regarding the Cited Paper's performance, utility, or flaws.

**CRITICAL RULES:**

1. **PERSPECTIVE (The "Golden Rule"):** 
   - Write comments **DIRECTLY** about the Cited Paper.
   - **FORBIDDEN PHRASES:** "The citing paper claims...", "The authors use this...", "In this work...", "Served as a foundation for..."
   - **REQUIRED FORMAT:** "The method demonstrates...", "The framework fails to...", "The algorithm is widely adopted for..."

2. **FILTERING (Neutrality is OK):**
   - If a citation is just a background mention or list (e.g., "Methods X, Y, Z exist"), mark type as "neutral".
   - **DO NOT** generate a review comment for neutral citations.

3. **SUMMARY TRANSLATION RULES (CRITICAL):**
   - When extracting insights from Research Questions (Q1-Q9), you **MUST REPHRASE** them into qualities of the Cited Paper.
   - **NEVER** output sentences starting with "Yes," or "No," in the summary.
   - **NEVER** output sentences containing "The citing paper" in the summary.
   
   **Translation Examples:**
   - *Bad (Q1 Input):* "Yes, the citing paper adopts the method for protein folding."
   - *Good (Summary Output):* "The method proves effective and adaptable for protein folding tasks."
   
   - *Bad (Q3 Input):* "Yes, the citing paper points out it is slow."
   - *Good (Summary Output):* "The approach suffers from significant computational latency."

**JSON OUTPUT:**
- Must be valid JSON.
- `type` enum: ["strength", "weakness", "neutral"].

You are an expert at identifying duplicate or highly similar comments in academic paper reviews.

Your task is to deduplicate a list of {category} comments by removing semantically similar ones.

**Rules:**
1. Keep the most comprehensive and well-articulated version when duplicates exist
2. Two comments are considered duplicates if they convey the same core insight, even with different wording
3. Return ONLY the deduplicated list in valid JSON format
4. Preserve the original phrasing of kept comments

**Output Format:**
{{
  "deduplicated_comments": [
    "comment 1",
    "comment 2",
    ...
  ],
  "removed_examples": [
    "example of removed comment 1",
    "example of removed comment 2",
    "example of removed comment 3"
  ],
  "removed_count": <number of duplicates removed>
}}
\end{MyVerbatim}
\end{tcolorbox}

\subsubsection{User Prompt}
\label{sec:1.2}
\begin{tcolorbox}[colback=yellow!10, colframe=black!50, title=Implicit Signal Enhancement -- User Prompt, breakable]
\begin{MyVerbatim}
### TASK: Review the "Cited Paper" based on the provided Citation Contexts.

---
### CITED PAPER INFORMATION:
- **Title:** {cited_title}
- **Citation Count:** {citation_count}
- **Abstract:** {cited_content}

---
### CITING PAPER INFORMATION:
- **Title:** {citing_title}
- **Content:** {citing_content}

---
### CITATION CONTEXTS (Evidence):
{contexts_list}

---

### INSTRUCTIONS FOR PROCESSING CITATIONS:

1. **Context Analysis**: Read the context to determine the Cited Paper's quality.
2. **Type Classification**:
   - **strength**: Explicit praise, adoption due to performance, or validation of robustness.
   - **weakness**: Explicit criticism, noted limitations, theoretical flaws, or failure cases.
   - **neutral**: Mere mention in related work, "we used X as a baseline" (without evaluation).
3. **Drafting Comments**:
   - If neutral: null or empty string.
   - If strength/weakness: Write purely about the Cited Paper (e.g., "The method achieves state-of-the-art results in...").

---

### RESEARCH QUESTIONS (Q1-Q9):

Please answer the following nine research questions based on the citation contexts provided:

**Q1:** Does the citing paper adopt or reuse the core method or algorithmic framework proposed by the cited paper?

**Q2:** Does the citing paper improve upon the cited paper's method? If yes, which parts are improved?

**Q3:** Does the citing paper point out any theoretical flaws or implementation limitations in the cited paper?

**Q4:** Does the citing paper claim that the cited paper's method generalizes poorly to specific tasks or datasets?

**Q5:** Does the citing paper use the same experimental datasets as the cited paper?

**Q6:** Does the citing paper introduce additional datasets on top of those used in the cited paper?

**Q7:** Does the citing paper reuse the experimental settings or evaluation metrics of the cited paper?

**Q8:** Does the citing paper introduce new metrics to complement or correct the evaluation used in the cited paper?

**Q9:** Does the citing paper argue that the cited paper overlooked key aspects such as efficiency, robustness, or generalization?

**Instructions for answering:**
- Answer based on evidence from the citation contexts.
- If the answer is "No" or "Not mentioned", simply state "No." and move on.
- Provide evidence quotes from the contexts to support your answers.

---

### SUMMARY GENERATION LOGIC (STRICT):

Construct the `summary` arrays using this logic:

**For `strengths`:**
1. Collect `review_style_comment` from citations where type="strength".
2. **Translate** positive Q1-Q9 answers:
   - If Q1 (Adoption) is Yes → Add: "The framework is robust enough to serve as a foundation for downstream tasks."
   - If Q2 (Improvement) is Yes → Add: "The method provides a strong baseline, though it leaves room for specific optimizations."
   - If Q5/Q7 (Re-use) is Yes → Add: "The experimental settings and metrics have become a standard benchmark."
3. **FILTER:** Remove any sentence starting with "Yes" or containing "citing paper".

**For `weaknesses`:**
1. Collect `review_style_comment` from citations where type="weakness".
2. **Translate** negative Q1-Q9 answers:
   - If Q3 (Flaws) is Yes → Add: "Theoretical or implementation flaws have been identified in the proposed approach."
   - If Q4 (Generalization) is Yes → Add: "The method demonstrates poor generalization on specific datasets."
   - If Q9 (Overlooked aspects) is Yes → Add: "Key aspects such as efficiency or robustness were insufficiently addressed."
3. **FILTER:** Remove any sentence starting with "No" or containing "citing paper".

---

### OUTPUT FORMAT:

Please provide your response in the following JSON format:

{
  "cited_paper": {
    "title": "{cited_title}",
    "content": "{cited_content}",
    "citation_count": "{citation_count}"
  },
  "citations": [
    {
      "id": 1,
      "citing_paper_title": "{citing_title}",
      "type": "strength | weakness | neutral",
      "context": "{merged_context}",
      "review_style_comment": "concise reviewer-style comment about cited paper only"
    }
    // ... more citations
  ],
  "review_reasoning": {
    "Q1": {
      "question": "Does the citing paper adopt or reuse the core method or algorithmic framework proposed by the cited paper?",
      "answer": "",
      "evidence": [""]
    },
    "Q2": {
      "question": "Does the citing paper improve upon the cited paper's method? If yes, which parts are improved?",
      "answer": "",
      "evidence": [""]
    },
    "Q3": {
      "question": "Does the citing paper point out any theoretical flaws or implementation limitations in the cited paper?",
      "answer": "",
      "evidence": [""]
    },
    "Q4": {
      "question": "Does the citing paper claim that the cited paper's method generalizes poorly to specific tasks or datasets?",
      "answer": "",
      "evidence": [""]
    },
    "Q5": {
      "question": "Does the citing paper use the same experimental datasets as the cited paper?",
      "answer": "",
      "evidence": [""]
    },
    "Q6": {
      "question": "Does the citing paper introduce additional datasets on top of those used in the cited paper?",
      "answer": "",
      "evidence": [""]
    },
    "Q7": {
      "question": "Does the citing paper reuse the experimental settings or evaluation metrics of the cited paper?",
      "answer": "",
      "evidence": [""]
    },
    "Q8": {
      "question": "Does the citing paper introduce new metrics to complement or correct the evaluation used in the cited paper?",
      "answer": "",
      "evidence": [""]
    },
    "Q9": {
      "question": "Does the citing paper argue that the cited paper overlooked key aspects such as efficiency, robustness, or generalization?",
      "answer": "",
      "evidence": [""]
    }
  },
  "summary": {
    "strengths": [
      "Comments from type='strength'",
      "Positive insights from Q1-Q9 (Only if Answer is Yes)"
    ],
    "weaknesses": [
      "Comments from type='weakness'",
      "Negative insights from Q1-Q9 (Only if Answer is Yes)"
    ]
  }
}

---

**ERROR PREVENTION GUIDELINES:**

1. **CONTEXT PRESERVATION:** 
   - NEVER modify the "context" field - use exactly what's provided
   - If context seems incomplete, still use it as-is

2. **ARRAY INTEGRITY:**
   - Ensure the "citations" array has exactly {context_count} items
   - Items must be in order (ID 1 to {context_count})
   - Do not add extra items or remove any

3. **TYPE SAFETY:**
   - "type" must be exactly "strength", "weakness", or "neutral"
   - If unsure between strength/weakness, choose based on dominant tone

4. **JSON VALIDATION:**
   - Test that your output is valid JSON before returning
   - Ensure all required fields are present
   - Escape special characters properly

**REMEMBER:** Complete processing of all {context_count} contexts is mandatory.

Here are the {category} comments to deduplicate:

{json.dumps(comments, indent=2, ensure_ascii=False)}

Please deduplicate these comments and return the result in the specified JSON format.
\end{MyVerbatim}
\end{tcolorbox}

\subsection{Prompts used in Training Data Construction with CoT}
\label{sec:2}

\subsubsection{Prompts used in Evidence Extraction}
\label{sec:2.1}
\begin{tcolorbox}[colback=yellow!10, colframe=black!50, title=Evidence Extraction Prompt, breakable]
\begin{MyVerbatim}
# Task:You are a paper analysis expert. Given a complete paper and a review comment about the paper, please find at least 1 core paragraph or sentence from the paper that directly supports this review comment.

    # Paper Content (complete)
    ```markdown
    {paper_content_full}
    ```

    # Review Comment
    Type: {claim_type}
    Content: {claim}

    # Output Requirements
    **Must output in pure JSON format with no markdown markers or extra text**

    {{
    "evidences": [
        {{
        "text": "Exact excerpt from original paper (50-200 words)",
        "reason": "How this excerpt supports the review comment (30-80 words)"
        }}
    ]
    }}

    Notes:
    1. "text" must be an exact copy from the paper
    2. Output 1-3 pieces of evidence
    3. Output JSON directly without ```json``` wrapper
    4. All content must be in English
\end{MyVerbatim}
\end{tcolorbox}

\subsubsection{Prompts used in Chain-of-Thought Synthesis}
\label{sec:2.2}

\paragraph{System Prompt}
\label{sec:2.2.1}
\begin{tcolorbox}[colback=yellow!10, colframe=black!50, title=Chain-of-Thought Synthesis -- System Prompt, breakable]
\begin{MyVerbatim}
You are a top-tier conference reviewer. Please conduct rigorous logical reasoning analysis based on the paper's original text and extracted evidence.

    # Reasoning Requirements
    1. Start DIRECTLY with quotes from the original text
    2. NEVER use phrases like: "The paper discusses", "The first evidence shows", "Evidence 1 states", "Furthermore, the second evidence"
    3. Structure: Direct quote → Brief analysis (1-2 sentences) → Next quote → Brief analysis → Conclusion
    4. Keep CONCISE: 2-4 sentences per evidence, 100-150 words total
    5. Quote format: "original text content"
    6. End with: "Therefore, this paper's {strength/weakness} is: [specific point]"

    # Output Format (GOOD):
    The paper states "quote A". This shows [brief insight]. Additionally, "quote B" demonstrates [brief analysis]. Therefore, this paper's weakness is: [specific conclusion in 1 sentence].

    # Output Format (BAD - avoid these):
    "The paper discusses in the first evidence..."
    "Furthermore, the second evidence states..."
    "Finally, the third evidence shows..."
    Long paragraphs with excessive explanation

    # Key Principles
    - CONCISE and DIRECT
    - NO meta-commentary about evidence structure
    - Maximum 150 words
    - Clear conclusion
    - All output in English
\end{MyVerbatim}
\end{tcolorbox}

\paragraph{User Prompt}
\label{sec:2.2.2}
\begin{tcolorbox}[colback=yellow!10, colframe=black!50, title=Chain-of-Thought Synthesis -- User Prompt, breakable]
\begin{MyVerbatim}
# Review Comment
    Type: {'Strength' if is_strength else 'Weakness'}
    Content: {claim}

    # Evidence from Paper
    {evidence_text}

    # Task
    Write a CONCISE analysis :
    1. Start with direct quotes - NO "The paper discusses" or "Evidence shows"
    2. Brief analysis after each quote (1-2 sentences)
    3. End with: "Therefore, this paper's {conclusion_prefix}: [specific point]"

    Keep it SHORT and DIRECT.
\end{MyVerbatim}
\end{tcolorbox}

\subsubsection{Prompts used in Multi-model Cross-validation (Faithfulness Audit)}
\label{sec:2.3}
\begin{tcolorbox}[colback=yellow!10, colframe=black!50, title=Multi-model Cross-validation Prompt, breakable]
\begin{MyVerbatim}
# Task
You are a rigorous scientific review expert. Please verify the faithfulness of the following reasoning chain.

# Original Paper Excerpts
{json.dumps(evidence_texts, ensure_ascii=False, indent=2)}

# Reasoning Chain
{reasoning}

# Verification Requirements
Please check the truthfulness and logical coherence of the reasoning item by item.

**Must output in pure JSON format with no markdown markers or extra text**

{{
  "citation_check": {{
    "all_valid": true,
    "invalid_citations": []
  }},
  "logic_score": {{
    "score": 9,
    "deduction_reasons": [],
    "comments": "Overall assessment"
  }},
  "overall_pass": true
}}

Notes: Output JSON directly without ```json``` wrapper. All content must be in English.
\end{MyVerbatim}
\end{tcolorbox}

\subsection{Prompts used in Experiments}
\label{sec:3}

\subsubsection{Prompts used in LLM-as-Judge}
\label{sec:3.1}
\begin{tcolorbox}[colback=yellow!10, colframe=black!50, title=LLM-as-Judge Prompt, breakable]
\begin{MyVerbatim}
You are an experienced Area Chair at top-tier academic conferences. Your task is to evaluate the quality of a peer review for a given paper based on specific evaluation criteria. Your assessment must be professional, objective, and well-justified.

You will receive an academic paper and its corresponding peer review.

First, carefully read and fully understand both the paper content and the review.

Then, evaluate the review quality based on the following dimensions:

1. Comprehensiveness: Does the review cover all important aspects of the paper, including but not limited to: importance of the research problem, innovation and originality, methodological rigor, experimental design and result analysis, potential impact on the field, and other critical aspects?


2. Specificity: Does the review address specific issues in this particular paper, rather than providing generic comments that could apply to any paper?

3. Evidence Support: Does the review support its observations and feedback by citing specific examples, sections, or experimental data from the paper? Are the citations faithful to the original paper?

4. Consistency: Is the review internally consistent in logic? Are there any contradictions or conflicting viewpoints?



Before providing any scores, analyze the review's performance on each evaluation dimension step by step.

For each dimension, provide a brief justification, then assign a score based on the following criteria:

* 0-2: Severely inadequate - Does not meet basic reviewing standards
* 3-4: Below acceptable - Requires major improvements
* 5-6: Acceptable - Meets minimum standards but has obvious limitations
* 7-8: Good - Exceeds typical expectations with only minor shortcomings
* 9-10: Excellent - Very high quality with minimal or no obvious flaws

Output your evaluation in the following JSON format:

{
  "Comprehensiveness": {
    "reason": "...",
    "score": 0
  },
  "Specificity": {
    "reason": "...",
    "score": 0
  },
  "Evidence Support": {
    "reason": "...",
    "score": 0
  },
  "Consistency": {
    "reason": "...",
    "score": 0
  }
}

Output ONLY the final JSON object without any additional text or explanation.
\end{MyVerbatim}
\end{tcolorbox}

\subsubsection{Prompts used in Prompt-based baselines}
\label{sec:3.2}
\begin{tcolorbox}[colback=yellow!10, colframe=black!50, title=Prompt-based Baseline Prompt, breakable]
\begin{MyVerbatim}
You are a top-tier conference reviewer. Please read the complete paper and conduct an in-depth analysis of the paper's strengths and weaknesses.\n\n# Analysis Requirements\n- Each strength/weakness must have rigorous reasoning process\n- Each key argument must cite the original text (format: \"quoted text\")\n- Reasoning must demonstrate causal logic\n- Use natural and fluent review language\n- All output must be in English\n\n# Output Format\nPlease structure your review as follows:\n\n# Strengths\n\n**S1.** [First strength with quote and analysis]\n\n**S2.** [Second strength with quote and analysis]\n\n...\n\n# Weaknesses\n\n**W1.** [First weakness with quote and analysis]\n\n**W2.** [Second weakness with quote and analysis]\n\n...\n\nEach point should follow this pattern:\nThe paper states \"[exact quote from paper]\". [Your analysis]. Therefore, this paper's strength/weakness is: [conclusion].
{full content of paper}
Please conduct a comprehensive review of this paper, analyzing its strengths and weaknesses.

\end{MyVerbatim}
\end{tcolorbox}

\subsubsection{Prompts used in Underlying Issue Matching}
\label{sec:3.3}
\begin{tcolorbox}[colback=yellow!10, colframe=black!50, title=Underlying Issue Matching Prompt, breakable]
\begin{MyVerbatim}
You are an expert in matching research issues. Your task is to determine whether two weaknesses from paper reviews point to the same underlying research issue.

## What is an "Underlying Research Issue"?

An underlying research issue refers to a **fundamental research limitation** in the study design and methodology, rather than specific wording or local implementation details. Here are several typical examples (not exhaustive):

1. **Inadequate Problem Modeling**: e.g., oversimplifying real-world problem complexity, ignoring important constraints
2. **Limited Applicability**: e.g., method only works in specific domains, has data scale requirements, limited generalization ability
3. **Missing Theoretical Analysis**: e.g., lack of convergence proof, insufficient complexity analysis

## Matching Criteria

Two weaknesses point to the same underlying research issue if and only if they criticize **the same fundamental research limitation**, even when:
- The wording is completely different
- They focus on different specific aspects (one discusses experiments, another discusses theory, but both point to the same fundamental flaw)
- The severity judgments differ

## Your Task

Human reviewer's weakness:
{human_weakness}

AI review system (EchoReviewer) weaknesses:
{chr(10).join([f"{i+1}. {w}" for i, w in enumerate(ai_weaknesses)])}

Please determine whether the human reviewer's weakness can find a corresponding item in the AI's weakness list that points to **the same underlying research issue**.

## Output Format

Please strictly follow this JSON format (do not include markdown code block markers):
{{
    "matched": true/false,
    "matched_index": <AI weakness number (1-based) if matched, otherwise null>,
    "reasoning": "<Brief explanation: if matched, explain what common underlying issue they both point to; if not matched, explain why no correspondence was found>"
}}

You are an expert in analyzing research paper weaknesses. Your task is to extract the **core underlying research issue** from a weakness description.

## Task

Given a weakness from a paper review, extract the **fundamental research problem** it points to. This should be:
- A concise phrase (3-8 words)
- Focus on the WHAT, not the HOW or WHY
- Generic enough to be recognized across different papers
- Specific enough to be meaningful

## Examples

Weakness: "The paper only evaluates on MNIST and CIFAR-10, which are relatively simple datasets. More diverse and challenging datasets should be included."
Core Issue: Limited dataset diversity

Weakness: "The authors do not provide convergence proofs or theoretical guarantees for their proposed algorithm."
Core Issue: Missing convergence analysis

Weakness: "The comparison baselines are outdated and do not include recent state-of-the-art methods from 2023."
Core Issue: Incomplete baseline comparison

Weakness: "The method assumes that all features are independent, which rarely holds in real-world applications."
Core Issue: Unrealistic independence assumption

## Your Task

Given weakness:
{weakness_text}

Extract the core underlying research issue as a concise phrase (3-8 words).

## Output Format

Respond with ONLY the phrase, no explanation, no JSON, no quotes. Just the phrase itself.
Example output:
Limited dataset diversity

You are an expert in clustering research issues. Your task is to group similar research issues together.

## Task

Given a list of research issues extracted from paper reviews, group them into clusters where each cluster represents the same or very similar underlying problem.

## Issues to Cluster

{issues_text}

## Instructions

1. Group issues that represent the same fundamental problem (e.g., "Limited dataset diversity", "Insufficient dataset variety", "Small number of datasets" should be in one group)
2. Choose the most representative phrase for each cluster
3. Return ONLY valid JSON format (no markdown blocks)

## Output Format

{{
    "cluster_name_1": [1, 3, 5],  // issue numbers that belong to this cluster
    "cluster_name_2": [2, 4],
    ...
}}

Where cluster names are the most representative phrases, and values are lists of issue numbers (1-indexed).

Example:
{{
    "Limited dataset diversity": [1, 3, 7],
    "Missing convergence proof": [2, 5],
    "Unclear methodology description": [4, 6]
}}

\end{MyVerbatim}
\end{tcolorbox}

\subsubsection{Prompts used in Time Span Experiments}
\label{sec:3.4}
\begin{tcolorbox}[colback=yellow!10, colframe=black!50, title=Time Span Experiment Prompt, breakable]
\begin{MyVerbatim}
"""# Role
You are an experienced reviewer at top-tier conferences. Your task is to compare two AI-generated peer reviews.

# Important Notes
- Your task is **relative comparison**, not absolute judgment
- Please select the "relatively better" one

The paper corresponding to these reviews is:
{paper_content}

# Task
Please compare Review A and Review B across the following 5 dimensions:

## 1. Comprehensiveness
**Comparison Question**: Which review covers more key points and is more comprehensive?
**Focus Areas**:
- Does it discuss multiple aspects (methodology, experiments, writing, related work, etc.)?
- Does it cover issues at different levels (high-level contributions, technical details, presentation issues, etc.)?


## 2. Specificity
**Comparison Question**: Which review provides more specific and actionable feedback?
**Focus Areas**:
- Are there vague statements (e.g., "the method is innovative", "experiments are insufficient")?
- Does it provide clear suggestions for improvement?
- Does it mention specific sections, figures, or tables?

## 3. Evidence Support
**Comparison Question**: Which review cites more specific content from the paper as evidence?
**Focus Areas**:
- Does it reference specific experimental results, data, or examples?
- Does the language indicate the reviewer carefully read the paper (rather than making generic comments)?

## 4. Consistency
**Comparison Question**: Which review has more coherent internal logic without contradictions?
**Focus Areas**:
- Are multiple points within the same dimension consistent with each other?
- Is the reasoning process logical?
- Are the evaluation criteria applied consistently throughout?

## 5. Overall
**Comparison Question**: If you were a PC member, which review would you prefer to reference?
**Explanation**: This is a comprehensive judgment considering practicality, credibility, helpfulness, etc., and does not need to strictly align with the previous dimensions.

---

# Decision Rules
- For each dimension, you must choose: **A** / **B**
- Even if the advantage is minimal, please try to make a judgment

---

# Output Format
Please strictly follow this JSON format (output JSON directly **without markdown code block markers**):

{{
  "Comprehensiveness": "A",
  "Clarity": "B",
  "Specificity": "A",
  "Evidence_Support": "A",
  "Consistency": "B",
  "Overall": "A",
  
  "reasoning": {{
    "Comprehensiveness": "Review A discusses three aspects: methodological innovation, experimental design, and writing quality, while Review B mainly focuses on methodology with narrower coverage.",
    "Clarity": "Review B uses a clear bullet-point structure with each weakness in a separate paragraph; Review A's presentation is more flowing but slightly verbose, with less prominent key points.",
    "Specificity": "Both reviews have comparable specificity. Review A mentions 'baseline comparisons in Table 2', while Review B references 'ablation study in Section 3.2', both showing some level of specificity.",
    "Evidence_Support": "Review A frequently cites specific sections and figures (e.g., 'Figure 3', 'Section 4.1'), while Review B has fewer citations and more general statements.",
    "Consistency": "Review A praises the experiments as comprehensive in Strengths but criticizes the lack of key ablations in Weaknesses, showing contradiction; Review B is more logically consistent.",
    "Overall": "Overall, Review A is superior in comprehensiveness and evidence support. Despite internal consistency issues, it provides more valuable reference material."
  }}
}}

---

# Review A
{review_a}

---

# Review B
{review_b}"""
\end{MyVerbatim}
\end{tcolorbox}

\newpage
\section{Case Study and Qualitative Analysis}
\label{sec:case-study}

To further illustrate the practical utility of \textbf{EchoReviewer-7B}, we conduct a qualitative case study on a representative paper concerning \emph{slot consistency in task-oriented natural language generation (NLG)}. As shown in Table~\ref{tab:case-study-strengths} and Table~\ref{tab:case-study-weaknesses}, we present the automated review generated by EchoReviewer-7B for this paper, which was not included in the training data. This analysis aims to demonstrate how a citation-context--driven training paradigm enables the model to perform deep technical synthesis and critical reasoning beyond surface-level summarization.

\begin{itemize}
\item \textbf{Precision in Empirical Grounding.}
A salient strength of EchoReviewer-7B lies in its ability to accurately ground high-level evaluative statements in concrete and fine-grained technical details. As evidenced in S1 and S4, the model does not merely acknowledge the paper’s contributions, but explicitly references Eq.~(2) to verify the formal definition of slot consistency and further decomposes the reinforcement learning objective by citing Eq.~(10)--(14). Such precise alignment between critique and the paper’s mathematical formulations ensures that the feedback is firmly anchored in the actual modeling assumptions and algorithmic structure.

\item \textbf{Identification of Methodological Confounding.}
The model further exhibits advanced critical reasoning by identifying potential confounding factors in the experimental design. In W1, EchoReviewer-7B isolates a key confounder by noting that the performance gains reported in Table~2 may largely stem from the retrieval component (KNN), rather than the core rectification network (IRN). This analysis requires an integrated understanding of the system architecture and the empirical evidence presented in both Table~2 and the ablation results in Table~4.

\item \textbf{Synthesis of Long-term Scientific Constraints.}
Leveraging the collective evaluative signals embedded in citation contexts, the generated review emphasizes long-term scientific considerations, including computational efficiency (W3), robustness to noisy supervision (W4), and sensitivity to hyperparameters (W5). These critiques go beyond immediate benchmark performance and instead focus on reproducibility, scalability, and stability, which are critical concerns for real-world deployment and follow-up research.

\item \textbf{Actionable Refinement Suggestions.}
Beyond identifying weaknesses, EchoReviewer-7B consistently proposes technically actionable refinement paths. For example, in W2, the model suggests introducing a value-level semantic accuracy metric to prevent models from superficially satisfying the ERR metric through slot-token insertion while producing incorrect surface realizations. This recommendation reflects a nuanced understanding of semantic alignment issues in task-oriented NLG.
\end{itemize}

\newpage
\begin{table}[h]
\centering
\caption{Strengths identified in the automated review generated by EchoReviewer-7B for a paper on slot-consistent task-oriented NLG.}
\label{tab:case-study-strengths}

\begin{tabular}{p{0.95\linewidth}}
\toprule
\centering \textbf{Strengths} \tabularnewline
\midrule

\textbf{S1.} The paper makes a clear and practically motivated contribution by explicitly formalizing *slot consistency* via Eq. (2) \(g(f(\mathbf{x}))=g(\mathbf{x})\), directly targeting missing/redundant slots highlighted in Table 1 (e.g., missing NAME, hallucinated AREA). This formalization tightly matches the evaluation metric ERR in Eq. (15), making the research question well-scoped and measurable. The design choice to treat slot tokens as a set via the extractor \(g\) (Eq. (1)) is simple yet effective for task-oriented NLG where DAs are largely flat. To further strengthen the contribution, the paper could discuss how this set-based constraint handles duplicates or repeated mentions (e.g., whether repeating a slot token is counted as redundancy and how that interacts with fluency).

\\
\midrule

\textbf{S2.} The Iterative Rectification Network (IRN) is a novel post-editing framework that is clearly specified: it iteratively rewrites a candidate \(\mathbf{y}^{(k)}\) into \(\mathbf{y}^{(k+1)}\) using the pointer rewriter in Eq. (4), with termination when \(g(\mathbf{y}^{(k)})=g(\mathbf{x})\). The pointer rewriter’s action space (copy \(c(i)\) vs. generate \(w\), Eq. (8)) is well aligned with the goal of preserving fluent spans while correcting slot tokens, and the case study in Table 6 concretely demonstrates multi-step correction (removing hallucinated PRICE first, then inserting missing AUDIO by iteration 3). The paper’s inclusion of both supervised initialization (Eq. (9)) and RL fine-tuning provides a coherent training pipeline. A useful extension would be to report the empirical distribution of “number of iterations to converge” across datasets (beyond the single example in Table 6) to quantify runtime/latency impact in deployment.

\\
\midrule

\textbf{S3.} The experience replay + data aggregation scheme (Algorithm 1) and retrieval-based bootstrapping (Algorithm 2) address a real data sparsity issue: genuine mistaken cases may be limited and biased, so the method explicitly enriches training with pseudo inconsistent samples within tolerance \(\epsilon=2\). The ablation study in Table 4 provides evidence that these components matter: removing bootstrapping drops Laptop BLEU from 0.537 to 0.464 and worsens SER/ERR from 0.29\% to 0.83\%, indicating the model is not simply learning from baseline mistakes alone. The iterative replenishing of \(C\) with IRN-generated hard cases (Algorithm 1, lines 14–19) is a sensible curriculum-like mechanism. To make this more actionable and reproducible, the paper should clarify buffer size, sampling ratios between \(C\) and \(\Omega\), and sensitivity to \(\epsilon\), since Algorithm 2 can be computationally heavy (looping over all \((\hat{\mathbf{x}},\hat{\mathbf{z}})\in T\)).

\\
\midrule

\textbf{S4.} The reinforcement learning objective is well motivated by the stated mismatch between cross-entropy training and ERR evaluation, and the paper provides an explicit reward decomposition: slot consistency \(r^{SC}\) (Eq. (10)), language model fluency \(r^{LM}\) (Eq. (11)), and distant supervision reward \(r^{DS}\) (Eq. (12)), combined in Eq. (13). Table 4 supports the claim that \(r^{SC}\) is essential: removing it increases SER from 0.29\% to 0.75\%, while removing \(r^{DS}\) also degrades SER to 0.66\%, validating the need for both correctness- and guidance-related signals. The use of BLEU(\(\mathbf{y},\mathbf{z}\)) as the REINFORCE baseline (Eq. (14)) is a concrete engineering detail that improves stability, as the authors note it works better than likelihood averaging. A concrete improvement would be to include variance reduction details (e.g., reward normalization, entropy regularization) and report training stability curves, since REINFORCE can be sensitive and the current evidence is mainly end-point metrics.

\\
\midrule

\textbf{S5.} The experimental evaluation is broad and includes strong baselines (HLSTM, SCLSTM, TGen, ARoA, RALSTM) across four datasets, with the main result table (Table 2) showing consistent gains in both BLEU and ERR (e.g., on Television: BLEU 0.559 vs. 0.541 for RALSTM, ERR 0.35\% vs. 0.63\%). The “IRN as an add-on” experiment (Table 3) is particularly convincing because it shows ERR reductions when IRN is attached to existing systems (e.g., HLSTM on Television: ERR reduced by 2.29\% absolute). Human evaluation (Table 5) further supports improvements in informativeness (e.g., RALSTM 4.63 → 4.86 with IRN), aligning with the paper’s core claim about slot consistency. To strengthen evidence, the paper should report confidence intervals and annotator agreement for Table 5, and clarify whether “informativeness” directly corresponds to slot correctness (missing/redundant) or includes other semantic criteria.

\\
\bottomrule
\end{tabular}
\end{table}

\begin{table}[h]
\centering
\caption{Weaknesses identified in the automated review generated by EchoReviewer-7B for a paper on slot-consistent task-oriented NLG.}
\label{tab:case-study-weaknesses}

\begin{tabular}{p{0.95\linewidth}}
\toprule
\centering \textbf{Weaknesses} \tabularnewline
\midrule

\textbf{W1.} The paper frames IRN as improving “general NLG systems,” but the strongest system-level result is reported as “IRN (+KNN)” (Table 2), making it unclear how much improvement comes from retrieval (KNN) versus rectification. Table 4 partially addresses this (w/o IRN: BLEU 0.414, SER 0.88\%), but this is only on Laptop and does not isolate KNN’s contribution under the same conditions as Table 2. A concrete fix is to add a consistent comparison: KNN-only vs. IRN-only post-editing on neural baselines vs. KNN+IRN across all four datasets, including latency. This would better justify the claim that IRN is broadly beneficial rather than the performance being driven by retrieval coverage.

\\
\midrule

\textbf{W2.} The slot consistency definition and ERR metric treat slot presence as set membership (Eq. (1)–(2), Eq. (15)), which may not detect *misalignment between slots and values* after lexicalization (e.g., swapping values across slots or incorrect value realization). The paper’s examples focus on missing/redundant slots (Table 1, Table 6), but do not evaluate value correctness post-lexicalization, despite lexicalization being a required final step (Section 2.1). An actionable improvement is to include a value-level semantic accuracy metric (e.g., exact match of slot-value pairs after lexicalization, or a semantic parser-based check) and report it alongside ERR. This would ensure the method does not “game” ERR by inserting correct slot tokens but producing incorrect surface values.

\\
\midrule

\textbf{W3.} Bootstrapping via retrieval (Algorithm 2) appears computationally expensive because it iterates over all entries in \(T\) for each sampled \((\mathbf{x},\mathbf{z})\) (lines 5–11), which could be prohibitive on larger datasets. The paper does not report runtime, memory, or complexity of generating \(\Omega\), nor how often bootstrapping is performed per epoch. A concrete suggestion is to replace the full scan with approximate nearest neighbor indexing over slot sets (e.g., hashing by slot signature) and report wall-clock training time with/without bootstrapping. Additionally, sensitivity experiments over \(\epsilon\) and sample size \(V\) would clarify whether performance depends on heavy bootstrapping or is robust to cheaper approximations.

\\
\midrule

\textbf{W4.} The distant supervision labels used for supervised warm-start (Section 4) are generated by “simple rules” and the paper itself notes semantic ambiguity and unequal token importance, but it does not quantify how noisy these inferred actions are. Since \(r^{DS}\) is also part of the RL reward (Eq. (12)–(13)), label noise could bias learning, especially for paraphrastic edits where copying vs. generating is not uniquely determined. A concrete improvement is to report an estimated label accuracy on a small manually annotated subset or provide an ablation that varies the weight \(\gamma^{DS}\) rather than fixing equal weights. This would make the role of distant supervision more interpretable and reduce risk that gains depend on brittle heuristics.

\\
\midrule

\textbf{W5.} The RL reward mixes heterogeneous scales (negative set difference count in Eq. (10) vs. negative perplexity in Eq. (11) vs. log-prob sum in Eq. (12)) but uses equal weights \(\gamma^{SC}=\gamma^{LM}=\gamma^{DS}\) without justification. Because perplexity magnitude can dominate, the equal-weight choice may not be optimal or stable across domains (e.g., Television vs. Restaurant). Table 4 shows removing \(r^{LM}\) changes SER and BLEU modestly, but it does not address reward scaling or weight sensitivity. An actionable fix is to normalize each reward (e.g., z-score per batch) or tune \(\gamma\) on validation sets and report sensitivity plots, which would improve reproducibility and confidence that the method generalizes.

\\
\bottomrule
\end{tabular}
\end{table}

\end{document}